%% file: main.tex
\documentclass[10pt,twocolumn,letterpaper]{article}
\usepackage[pagenumbers]{cvpr}
\usepackage{amsmath,amssymb,booktabs,microtype,placeins,needspace}
\usepackage{tikz}
\usetikzlibrary{arrows.meta,positioning}
\input{writing-tools}

\usepackage[pagebackref,hidelinks]{hyperref}
\newcommand{\paperchapter}[1]{\section{#1}}
\newcommand{\papersection}[1]{\subsection{#1}}
\title{ContextHands: Completing Hand Motion\\
with Learned Context and Joint-Specific Body Kinematics}
\author{Taewoo~Kim\qquad Seunghun~Koh\qquad Hyosup~Shin\qquad Sungho~Jo\textsuperscript{\textdagger}\\[3pt]
{\normalsize Korea Advanced Institute of Science and Technology (KAIST)}\\[1pt]
{\small\textsuperscript{\textdagger}\,Corresponding author}}
\hypersetup{pdftitle={ContextHands: Completing Hand Motion with Learned Context and Joint-Specific Body Kinematics},pdfauthor={Taewoo Kim, Seunghun Koh, Hyosup Shin, and Sungho Jo}}
\newcommand{\contexthandstitleteaser}{}
\makeatletter
\g@addto@macro\@maketitle{\vspace{-18pt}\input{figures/teaser}}
\makeatother
\AddToHookNext{shipout/foreground}{%
  \put(0,-765){\makebox[\paperwidth][c]{%
    \shortstack[c]{\footnotesize This work has been submitted to the IEEE for possible publication.\\%
      \footnotesize Copyright may be transferred without notice, after which this version may no longer be accessible.}}}}
\makeatletter
\newcommand{\mainpaperspacing}{%
  \renewcommand\paragraph{\@startsection{paragraph}{4}{\z@}{1.5ex plus .4ex minus .2ex}{-1em}{\normalfont\normalsize\bfseries}}%
  \def\cvprsection{\@startsection{section}{1}{\z@}{-7pt plus -2pt minus -2pt}{3pt}{\large\bf}}%
  \def\cvprsubsection{\@startsection{subsection}{2}{\z@}{-5pt plus -2pt minus -2pt}{2.5pt}{\elvbf}}%
  \setlength{\textfloatsep}{8pt plus 2pt minus 2pt}%
  \setlength{\dbltextfloatsep}{8pt plus 2pt minus 2pt}%
  \setlength{\abovedisplayskip}{6pt plus 2pt minus 2pt}%
  \setlength{\belowdisplayskip}{6pt plus 2pt minus 2pt}%
  \widowpenalty=3500\relax\clubpenalty=3500\relax}
\makeatother
\begin{document}
\raggedbottom
\maketitle
\begingroup\mainpaperspacing
\begin{abstract}\input{cvpr/00_abstract}\end{abstract}
\input{cvpr/body}

\FloatBarrier
\clearpage
\endgroup
\noindent\textbf{Figure credits.}
The reference motions and object poses in our figures are adapted from
\href{https://huggingface.co/datasets/kelvin34501/OakInk-v2}{OakInk2}~\cite{zhan2024oakink2}.
The OakInk2-derived material in these figures is shared under
\href{https://creativecommons.org/licenses/by-sa/4.0/}{CC BY-SA 4.0}.
We rendered the body and hands using
\href{https://smpl-x.is.tuebingen.mpg.de/bodylicense.html}{SMPL-X Body}~\cite{pavlakos2019smplx}
(\href{https://creativecommons.org/licenses/by/4.0/}{CC BY 4.0}), courtesy of
the Max Planck Institute for Intelligent Systems. We adjusted materials,
lighting, camera views, alignment, and layout.

\input{cvpr/llm_usage_statement}
{\small\input{references.tex}}
\end{document}

%% file: writing-tools.tex
\usepackage{xcolor,graphicx}

\newcommand{\scoreci}[2]{#1\textsuperscript{\mdseries\unboldmath\color{black}\ensuremath{\pm}#2}}
\newcommand{\rankfirst}[1]{\textcolor{bestred}{{\bfseries\boldmath #1}}}
\newcommand{\ranksecond}[1]{\textcolor{secondblue}{{\bfseries\boldmath #1}}}

%% file: figures/teaser.tex
\ifdefined\contexthandstitleteaser
\noindent\begin{minipage}{\textwidth}\captionsetup{type=figure}
\else\begin{figure*}[t]
\fi
\centering
\includegraphics[width=\textwidth]{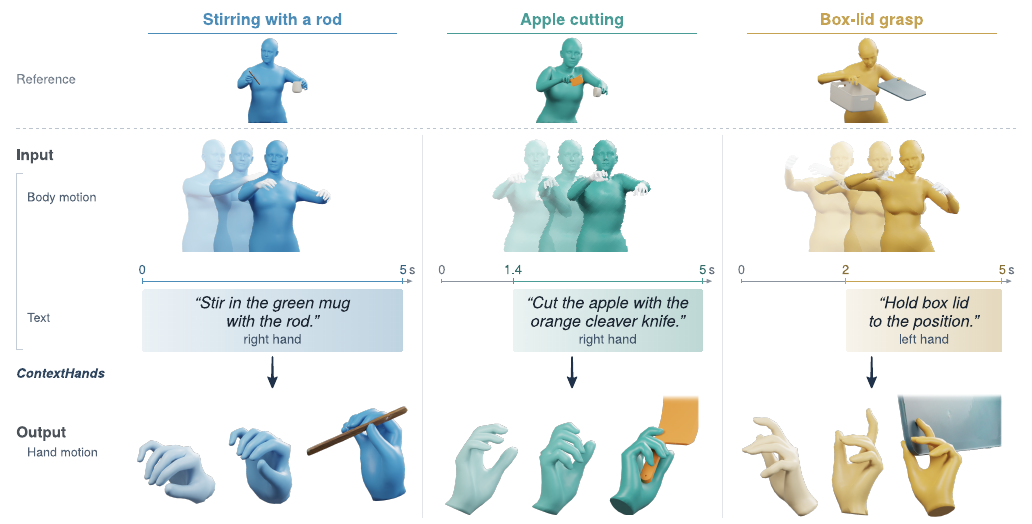}
\caption{ContextHands completes hand motion from input body motion and text descriptions assigned to individual hands and time intervals. Our method combines context learned from these inputs with joint-specific body kinematics to generate detailed hand articulation while preserving the body motion.}

\label{fig:teaser}
\ifdefined\contexthandstitleteaser\end{minipage}\par\vspace{8pt}
\else\end{figure*}\fi

%% file: cvpr/00_abstract.tex
Detailed hand motion contributes to the realism and expressiveness of generated human motion. Despite advances in human motion generation, synthesizing fine hand movements remains challenging. We focus on generating accurate hand motion during object manipulation. We introduce ContextHands, a diffusion model that completes hand motion from a body-motion sequence and text descriptions assigned to hands and time intervals. ContextHands learns a probabilistic context for each hand from the descriptions and body motion through auxiliary hand prediction and joint training with the generator. The joints of each hand share this context, while each joint queries spatial and temporal body features for its own kinematic condition. A single diffusion generator combines the shared context with these joint-specific conditions to produce detailed hand motion. On the OakInk2 dataset, ContextHands achieves lower hand joint position errors and lower Fréchet Hand Distance than all five methods adapted to this task. In qualitative comparisons, it more closely reproduces the hand poses used for grasping and tool use. These results demonstrate the effectiveness of our architecture for precise hand-motion generation from body motion and text. We will release our code and pretrained models to support further research.

%% file: cvpr/body.tex
\input{cvpr/01_introduction}
\input{cvpr/02_related_work}
\input{cvpr/03_method}
\input{cvpr/04_experiments}
\input{cvpr/05_results}
\input{cvpr/07_discussion}
\input{cvpr/06_conclusion}

%% file: cvpr/01_introduction.tex
\paperchapter{Introduction}
\label{sec:intro}
\ifdefined\contexthandstitleteaser\else\input{figures/teaser}\fi
Detailed hand motion is important for expressive and realistic human
motion. Hands convey intent through gestures and play a central role
in everyday object manipulation
\cite{ng2021body2hands,shi2026priorfirst,zhang2026handx}.
Despite progress in human motion generation, generating detailed hand
motion remains challenging~\cite{zhang2026handx}. Hands have many coupled
joints, and an action can be performed with different hand
configurations~\cite{shi2026priorfirst,qi2023bhd}.
Hand-motion completion generates these hand details to accompany an
already available body sequence.
One potential application is augmenting existing body-motion datasets
with hand motion guided by descriptions assigned to hands and time intervals.
This is particularly relevant because high-quality motion data that capture
both body and detailed hand motion remain limited~\cite{zhang2026handx}.

Body-conditioned methods exploit correlations between body and hand
motion~\cite{ng2021body2hands,qi2023bhd}.
However, body motion alone does not uniquely determine hand motion:
similar arm movements can accompany different finger
configurations~\cite{qi2023bhd}. Language provides complementary information about
actions, body-part configurations, and spatial
relations~\cite{delmas2022posescript,zhang2024both2hands}.
Together, body motion and language provide context for the hand motion
that accompanies an action.

This context can differ between hands and change over time: one hand
may remain open while the other closes and later opens again.
Descriptions therefore need not apply to both hands throughout a clip. OakInk2~\cite{zhan2024oakink2} provides descriptions of
manipulation events with hand-specific temporal annotations. We use
these existing descriptions and their hand--time assignments as inputs
for completion from body motion. This raises the question of how to use
body motion and the text descriptions to generate detailed hand motion.

BOTH2Hands~\cite{zhang2024both2hands} learns separate body-to-hand and
text-to-hand generators, then blends their predictions, while Prior-First,
Condition-Second~\cite{shi2026priorfirst} learns semantic adapters on a
frozen body--hand prior. We investigate a different conditioning design
based on two complementary uses of body motion: learning hand-level
context together with text, and providing kinematic conditions for
individual hand joints. We hypothesize that separating these roles within
a jointly trained model can improve hand-motion completion.
In ContextHands, a Hand Context Encoder learns these context features
from the text descriptions and body motion. Auxiliary hand prediction
links the features to hand articulation, while joint training with the
generator shapes them for the final completion task. We model the features
probabilistically because body motion and language still permit
different hand articulations~\cite{sohn2015cvae,petrovich2022temos}.

At each frame, the context feature is shared across the joints of a hand.
To generate detailed hand motion consistent with the body, we combine this
shared context with body conditions provided separately to each hand joint.
This allows individual joints to draw on different body cues while using
the same hand-level context. The Joint-Query Kinematic Encoder preserves
joint and anatomical-part features across multiple temporal scales; each
hand joint queries them for its own condition. A single Hand Diffusion
Generator combines these conditions with the shared hand context to
generate rotations while keeping the input body motion, including wrist
positions, fixed.

We focus our evaluation on hand motion during object manipulation, where
detailed hand poses and movements are needed to represent how objects
are grasped and used~\cite{zhan2024oakink2}. We assess both reconstruction
accuracy and similarity to the distribution of real hand motion.
We evaluate ContextHands on OakInk2 and examine both conditioning paths
and joint versus staged training through ablations. Our contributions are:
\begin{itemize}
\item We propose ContextHands, a diffusion model that uses body motion in
two complementary roles: learning hand-level context together with text
and providing joint-specific kinematic conditions. These representations
are learned jointly with a single generator to produce detailed hand motion.
\item We evaluate hand-motion completion from body motion and text using
motion sequences involving object manipulation. The results demonstrate
the effectiveness of our design, with more accurate hand motion and a
closer match to the distribution of real hand motion than the compared methods.
\end{itemize}
% Distribution narrowing and explicit role inference are not measured claims.
% The deterministic control tests a training bundle, not sampling alone.

%% file: cvpr/02_related_work.tex
\paperchapter{Related Work}
\label{sec:related}

\paragraph{Hand motion generation.}
Several methods generate hand motion by learning correlations between
body and hand motion. Body2Hands~\cite{ng2021body2hands} predicts
conversational gestures from upper-body motion, while BHD~\cite{qi2023bhd}
uses separate hand branches and spatial and temporal memories to model
body--hand relationships. Language provides additional cues, but methods
differ in how they combine these conditions.
BOTH2Hands~\cite{zhang2024both2hands} first trains body- and
text-conditioned generators separately, then blends their predictions
through cross-attention. Prior-First,
Condition-Second~\cite{shi2026priorfirst} instead learns semantic
adapters on a frozen autoregressive body--hand prior with
kinematic-chain attention. ContextHands learns body-and-text context
and joint-specific body conditions jointly with a single hand generator.
Our hand-joint queries are computed from the input body motion before
denoising, whereas Prior-First derives queries for its kinematic attention
from noisy hand states.
Among dedicated text-to-hand models, HandMDM~\cite{bensabath2026handmdm}
uses descriptions derived from sign-language data, whereas
CLUTCH~\cite{thambiraja2026clutch} learns a hand-motion tokenizer and
language model from in-the-wild motion--text pairs.
HandX~\cite{zhang2026handx} distinguishes left-hand, right-hand, and
inter-hand descriptions and also supports wrist-trajectory constraints.
Other methods jointly generate body, hand, and facial motion
from speech~\cite{yi2023talkshow,liu2024emage} or
text~\cite{lu2024humantomato}.

\paragraph{Motion representations and generation.}
Motion generation from text uses diffusion
models~\cite{tevet2023mdm,zhang2024motiondiffuse,chen2023mld,zhang2023remodiffuse}
as well as autoregressive or masked prediction of discrete motion
tokens~\cite{zhang2023t2mgpt,jiang2023motiongpt,guo2024momask,pinyoanuntapong2024mmm}.
A related question is how learned representations connect language to
motion. TEMOS~\cite{petrovich2022temos} links text and motion latent
distributions through a shared motion decoder, whereas
MotionCLIP~\cite{tevet2022motionclip} aligns a motion autoencoder with
the CLIP embedding space~\cite{radford2021clip}.
TMR~\cite{petrovich2023tmr} combines motion reconstruction with contrastive text--motion alignment to learn representations; MoLingo~\cite{he2026molingo} uses frame-level language
supervision to learn motion representations.

Hand-specific supervision can also guide text-driven generation:
TSHaMo~\cite{cheng2026tshamo} transfers guidance from a teacher with
auxiliary hand inputs during training to a student that uses only text
at inference.

\paragraph{Controllable motion and interaction synthesis.}
Control beyond a sequence-level description can specify where or when
motion occurs. Spatial constraints guide diffusion in
GMD~\cite{karunratanakul2023gmd} and
OmniControl~\cite{xie2024omnicontrol}, while
FineMoGen~\cite{zhang2023finemogen} uses spatiotemporal descriptions.
KinMo~\cite{zhang2025kinmo} structures language--motion
alignment around global actions, kinematic groups, and their
interactions. Complementary work composes motion across time or body
parts~\cite{athanasiou2022teach,athanasiou2023sinc,petrovich2024stmc}.
Interaction synthesis models relationships between the two hands or between hands and objects. InterHandGen~\cite{lee2024interhandgen} and
HandDiffuse~\cite{lin2025handdiffuse} model two-hand interactions.
Text-conditioned object interaction includes full-body synthesis in
IMoS~\cite{ghosh2023imos} as well as hand-focused generation.
For the latter, DiffH2O~\cite{christen2024diffh2o} separates grasping
from subsequent manipulation, while Text2HOI~\cite{cha2024text2hoi}
predicts contact from text and object geometry before generating motion.
OakInk2's Task-aware Motion Fulfillment (TaMF)
baseline~\cite{zhan2024oakink2} pairs task descriptions with
hand-specific interaction segments and generates hand motion from
object trajectories and geometry. ContextHands instead completes hand
motion from body motion and provided descriptions, without
object geometry at inference or explicit contact modeling.

%% file: cvpr/03_method.tex
\linespread{0.965}\selectfont
\paperchapter{Method}
\label{sec:method}
\ifdefined\contexthandsreading\else\input{figures/textbody_architecture}\fi
\papersection{Task definition}
\label{sec:task}
Given body motion $B$ over $T$ frames and descriptions $\mathcal T$, we
generate hand-joint rotations while keeping the body motion,
including wrist positions, fixed. Assignments $A$ specify the hand and
time interval for each description. These inputs are available during
training and inference; target hand motion is available only during training.

Following Prior-First~\cite{shi2026priorfirst}, we separate the given body
kinematics from the rotations to be generated. We model
$p(H\mid B,\mathcal T,A)$, where
$H\in\mathbb R^{T\times2\times16\times6}$ contains one wrist and 15
finger-joint rotations per hand in continuous 6D
form~\cite{zhou2019rotation}. These are local rotations relative to each
joint's parent in SMPL-X~\cite{pavlakos2019smplx}; wrist rotations are
relative to the forearm. We mask wrist-rotation features and exclude finger rotations from the body inputs. Forward kinematics combines the generated
rotations with the given body motion to obtain hand geometry.

\ifdefined\contexthandsreading\input{figures/textbody_architecture}\fi
\papersection{Method overview}
\label{sec:methodoverview}
ContextHands combines context shared within each hand with body cues
specific to its individual joints (Fig.~\ref{fig:textbodyarchitecture}).
The Hand Context Encoder (HCE) learns framewise features
$U_{t,h}\in\mathbb R^{96}$ for hand $h$, guided by auxiliary hand prediction.
The Joint-Query Kinematic Encoder (JQKE) reads body information separately
for each hand joint to obtain $C(B)$. A single Hand Diffusion
Generator uses both conditions to complete the hand motion.

Let $c=c(B,\mathcal T,A)$ denote the encoded text--body context.
The generative process is
\begin{equation}
\begin{aligned}
 z &\sim p_\phi(z\mid c),\\
 U &= D_\eta(c,z),\\
 H &\sim p_\theta(H\mid C(B),U).
\end{aligned}
\label{eq:textbodydraft}
\end{equation}
Here, $p_\phi$, $D_\eta$, and $p_\theta$ are the conditional prior,
feature decoder, and diffusion generator. The two paths have separate body encoders and are
trained jointly with the generator.

\ifdefined\contexthandsreading\else\Needspace{6\baselineskip}\fi
\papersection{Hand Context Encoder}
To follow descriptions assigned to different hands and intervals, the
Hand Context Encoder maintains a feature for each hand and frame.
We use a Transformer~\cite{vaswani2017attention} to extract a body-motion feature
for each hand at each frame. For each description, a learned attention layer pools
token features from frozen DistilBERT~\cite{sanh2019distilbert} into a single vector.
The assignments $A$
place each description feature at its corresponding hand and frames,
with zero text features where descriptions are absent. Adding the body
and text features gives context $c$ (Fig.~\ref{fig:textbodyarchitecture}).

Body motion and text provide cues about the action while leaving some details
of finger pose unspecified. We model this variation using a conditional
prior~\cite{sohn2015cvae}. Because the two hands can play different roles in
the same action, we sample a separate latent vector $z_h$ for each hand.
Some aspects of hand pose, such as the degree of finger flexion, can persist
as the arm moves. We therefore share $z_h$ across that hand's frames so that
a single sample can influence its pose throughout the motion. To obtain the
prior $p_\phi(z\mid c)$, a learned query for each hand summarizes $c$ across
both hands and all valid frames. From this summary, HCE predicts the mean
and variance of each coordinate of a Gaussian latent vector. At each
frame $t$, the feature decoder adds a projection of the sampled $z_h$ to
$c_{t,h}$. Each frame thus receives the same hand-specific sample alongside
its own body and text cues. The decoder then applies attention across hands
and frames to produce the context features $U$.

During training, the target hand motion $H$ is also available. We use $H$
together with the body and text context $c$ to estimate an approximate
posterior $q_\psi(z\mid c,H)$. A KL loss encourages the posterior and the
conditional prior to remain close. We draw samples $z^p$ and $z^q$ from
the prior and posterior, respectively, and pass each through the same
feature decoder to obtain
\begin{equation}
 U^p=D_\eta(c,z^p),\qquad U^q=D_\eta(c,z^q).
 \label{eq:hcepasses}
\end{equation}
The same generator then predicts the target hand motion in two passes, one
conditioned on $U^p$ and the other on $U^q$. Both passes use the same
noisy hand input, diffusion step, and body condition $C(B)$. The generator
thus learns to use both the features informed by the target motion and
those available from body motion and text alone.

An auxiliary prediction task encourages the context features to retain
information about hand pose. We use principal component analysis (PCA) to
represent the joint rotations of each hand at each frame with 16 coefficients.
We fit a separate PCA basis to each hand's training rotations and keep both
bases fixed. A shared linear layer predicts the 16 standardized coefficients
from $U^q$. We minimize the mean squared error between these predictions and
the coefficients computed from the target pose. This objective trains the
shared feature decoder through $U^q$. At inference, the same decoder takes
a sample from the conditional prior and produces $U^p$, which is passed
directly to the generator.

\papersection{Joint-Query Kinematic Encoder}
\label{sec:bodyconditioning}
HCE provides a shared context for the joints of each hand. We give each
hand joint its own query in JQKE so that joints within the same hand can
draw on different aspects of the same body motion. Each query has access
to individual body-joint features as well as summaries of body regions
and motion over different time spans.

Spatial attention exchanges information among body joints within each
frame, with learned biases for anatomical relationships. To capture
motion over different time ranges, three temporal branches use
short-range convolution, dilated convolutions, and sequence-wide
attention. Each retains a feature for every joint and frame. For each
body joint at each frame, we use its spatial feature to weight the
outputs of the three temporal branches. We project the weighted sum and
add it back to the spatial feature to obtain $F_t$.

To make the motion of a body region such as an arm available as a single
feature, we pool the joint features in $F_t$ within each anatomical
group using learned weights. This gives the part summaries $P_t$.
We also retain a separate summary from each temporal branch so that a
query can draw directly on a particular time scale. To form a branch's
summary at frame $t$, we average its 22 joint features at that frame and
project the result. The three summaries form $S_t$. For each
$t\in\{1,\ldots,T\}$, the resulting feature bank is
\begin{equation}
 E_t=[\widetilde F_t;\widetilde P_t;\widetilde S_t]
 \in\mathbb R^{30\times256},
 \label{eq:bodybank}
\end{equation}
containing 22 joint features, five part summaries, and three scale
summaries. Tildes denote the addition of learned type embeddings.

Each hand-joint query combines its joint and side identity with same-side
wrist and arm features and a whole-body summary. At frame $t$, all 32
hand-joint queries in $Q_t$ read the same 30 body features in $E_t$,
including features from the opposite side of the body. Two residual
cross-attention blocks update each query by weighting these features
separately for each hand joint:
\begin{equation}
 C_t=f_{\mathrm{CA}}(Q_t,E_t)
 \in\mathbb R^{32\times256}.
 \label{eq:handconditioning}
\end{equation}
The updated query for each joint serves as its body condition in the
generator. Collecting these conditions
over all $T$ frames gives $C(B)\in\mathbb R^{T\times32\times256}$.
The queries and bank depend only on $B$, so $C(B)$ is computed once
before diffusion sampling.

\papersection{Hand Diffusion Generator}

At each diffusion step, the generator estimates clean hand-joint rotations
from a noisy motion sequence using the hand context from HCE and the
joint-specific body conditions from JQKE (Fig.~\ref{fig:textbodyarchitecture},
right). During training, we standardize the target rotations using
training-set statistics before adding noise. We map each noisy 6D rotation
to a 256-dimensional feature vector, or joint token, and add embeddings
that identify the joint, hand side, and frame. Before the first block, we
project the hand context $U_{t,h}$ to 256 dimensions and add it to all 16
joint tokens of hand $h$ at frame $t$. The joints of each hand thus share
the same context at a given frame.

Each of the eight denoiser blocks updates every joint token using its own
body condition. For joint $j$ at frame $t$, we add an embedding of
diffusion step $s$ to the body condition $C_{t,j}$. An MLP takes this
sum and the normalized joint token as input and predicts an update,
which we add to the token. Temporal self-attention then combines
information from the same joint across frames, and within-hand
self-attention combines information from the joints of the same hand
within each frame. Each block ends with a feed-forward update. After the
eighth block, joint-specific output heads predict the clean rotations
$\hat{H}_0$. As in MDM~\cite{tevet2023mdm}, we apply forward kinematics
to the predicted rotations and supervise the resulting hand positions.

We use a 100-step cosine noise schedule~\cite{nichol2021improved}.
At inference, DDPM~\cite{ho2020ddpm} starts from Gaussian noise sampled
independently of the HCE latent. At each reverse step, it runs the
generator once and uses the predicted clean rotations to update the motion
sequence. The hand context $U$ and body conditions $C(B)$ remain fixed
throughout sampling.

\papersection{Training}
\label{sec:training}
Given the same noisy input, the generator predicts the target hand motion once
with the prior context $U^p$ and once with the posterior context $U^q$.
Following Sohn et al.~\cite{sohn2015cvae}, we minimize
\begin{equation}
\begin{aligned}
\mathcal L={}&\alpha\mathcal L_{\mathrm{hand}}^{p}
 +(1-\alpha)\mathcal L_{\mathrm{hand}}^{q}\\
 &+0.02\,\mathcal L_{\mathrm{coef}}
 +\beta\overline D_{\mathrm{KL}}(q_\psi\Vert p_\phi).
\end{aligned}
\label{eq:textbodytraining}
\end{equation}
Both predictions use the same hand-motion loss,
\begin{equation}
 \mathcal L_{\mathrm{hand}} =
 \mathcal L_{\mathrm{6D}} + 0.1\,\mathcal L_{\mathrm{rot}}
 + \mathcal L_{\mathrm{FK}} + 0.05\,\mathcal L_{\mathrm{inc}}
 + 0.01\,\mathcal L_{\mathrm{acc}} .
 \label{eq:handmotionloss}
\end{equation}
The first three terms measure mean squared error (MSE) in 6D rotations,
geodesic error in local rotations, and coordinatewise L1 error in hand
positions obtained by forward kinematics.
The last two compare relative rotations by geodesic distance and changes in
successive rotation increments by L1 error. $\mathcal L_{\mathrm{coef}}$ is the
MSE between the standardized PCA coefficients predicted from $U^q$ and the
target coefficients. We increase the KL weight $\beta$ linearly during warm-up.

Since inference uses the prior, we adjust $\alpha$ using the difference in
prior-conditioned loss on separate training recordings after two trial
updates from the same model state~\cite{ren2018reweight}. We restore the model
and optimizer before the actual update.
\par\linespread{1}\selectfont

%% file: figures/textbody_architecture.tex
\ifdefined\contexthandsreading\begin{figure}[tbp]\else\begin{figure*}[t]\fi
\centering
\includegraphics[width=\textwidth]{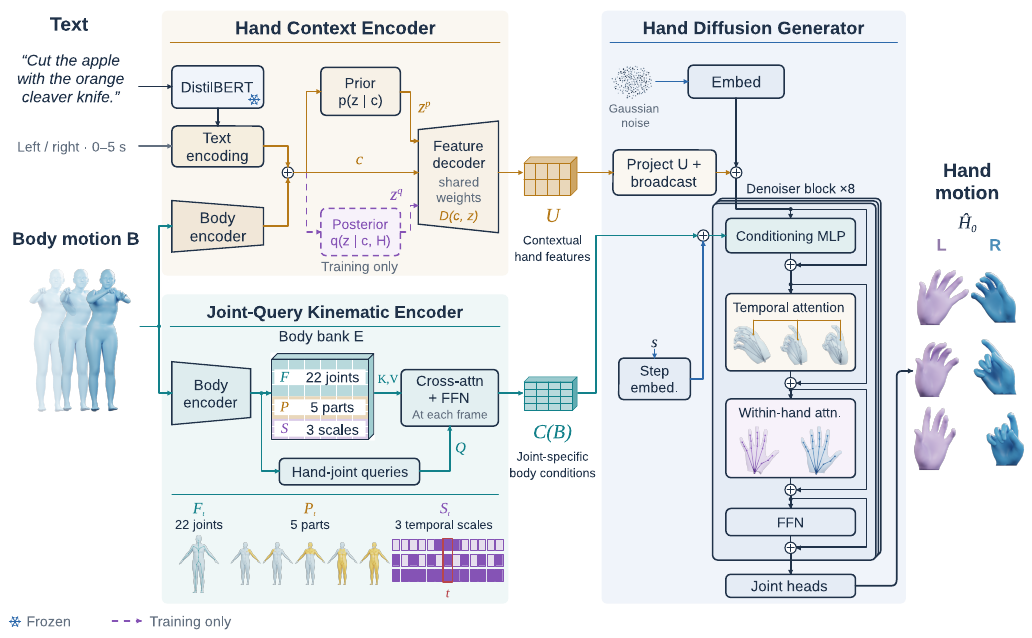}
\caption{Overview of ContextHands. From body motion and text descriptions assigned to hands and time intervals, the Hand Context Encoder produces contextual features $U$ shared within each hand. The Joint-Query Kinematic Encoder provides joint-specific body conditions $C(B)$. The Hand Diffusion Generator combines both conditions to generate hand-joint rotations. Prior and posterior features are processed in separate passes during training; inference uses the prior.}
\label{fig:textbodyarchitecture}
\ifdefined\contexthandsreading\end{figure}\else\end{figure*}\fi

%% file: cvpr/04_experiments.tex
\paperchapter{Experimental Setup}
\label{sec:experiments}
\input{tables/textbody_comparison}
\paragraph{Dataset.}
We evaluate hand-motion completion on OakInk2~\cite{zhan2024oakink2}, which
captures body and hand motion together during object manipulation.
All methods receive the full body sequence and descriptions assigned to hands and
time intervals. We compare predictions with the corresponding hand motion and
retain clips with partial or no text coverage. We split the official training set
by recording for training and validation and use the official test set.
We downsample from 120 to 30\,FPS and form non-overlapping clips of up to 150 frames (5 seconds).

\paragraph{Baselines.}
We compare with MDM~\cite{tevet2023mdm},
MLD~\cite{chen2023mld}, T2M-GPT~\cite{zhang2023t2mgpt},
BOTH2Hands~\cite{zhang2024both2hands}, and the diffusion variant of
HandX~\cite{zhang2026handx}. These baselines were designed for different motion generation tasks
and use different conditioning inputs. To compare them on our task, we adapt
each to generate hand-joint rotations from the same body motion and text
descriptions, with explicit hand and time assignments. We adjust the input and
output representations and validity masks while retaining each method's core
architecture, pretrained text encoder, and original training and sampling
procedures wherever compatible. HandX uses T5-base~\cite{raffel2020t5}, while the other baselines use
CLIP~\cite{radford2021clip}. All motion networks are trained from
scratch on the same OakInk2 training split. Adapted methods are marked
with *.

Prior-First~\cite{shi2026priorfirst} is not included in our OakInk2
comparison because its training code was unavailable when we conducted
the experiments.

\paragraph{Implementation details.}
All methods use a 24-hour training budget on a single RTX~4090.
We train ContextHands with AdamW~\cite{loshchilov2019adamw}.
Following the validation-FID selection used by T2M-GPT~\cite{zhang2023t2mgpt},
we select each method's final-stage checkpoint with the lowest validation FHD
within this budget. All test metrics use this checkpoint and fixed inference settings.

\paragraph{Evaluation metrics.}
We generate sixteen sequences per test input ($K=16$) from each selected
checkpoint. For each sequence, we compute the mean per-joint position error
(MPJPE) in millimeters over 42 hand points, including the two given wrist
positions. For each clip, we report the minimum (\mbox{MinMPJPE}) and mean
(\mbox{AvgMPJPE}) of the 16 sequence-level errors, following prior work on
evaluating multiple generated motions~\cite{petrovich2022temos}.
Hand positions are obtained by combining generated rotations with
the given body motion through forward kinematics, without rotation
or scale alignment. Both metrics weight recordings equally.
Motivated by the wrist--forearm
evaluation in Prior-First~\cite{shi2026priorfirst}, we also report
wrist rotation error as the mean SO(3) geodesic distance between
predicted and reference orientations, in degrees~\cite{zhou2019rotation}.

We measure distributional similarity with
Fr\'echet Hand Distance
(FHD)~\cite{qi2023bhd}, which compares generated and reference feature
distributions using the Fr\'echet Inception Distance (FID)
formulation~\cite{heusel2017fid}.
We extract features from a
frozen hand-motion encoder adapted from a gesture
autoencoder~\cite{yoon2020speech}, trained and selected using only recordings from the generator's training split. In the same feature space, Diversity
measures pairwise distances across test-set motion windows, with
GT diversity as a reference. Multimodality (MModality) measures
variation among completions of the same input through pairwise
feature distances~\cite{chen2023mld,zhang2023t2mgpt}. All three
feature-based metrics use the same set of 64-frame windows.

Table~\ref{tab:textbodycomparison} reports means and approximate 95\% confidence
intervals across generation repeats at fixed checkpoints. We report point estimates
for \mbox{MinMPJPE} and Multimodality, which use all sixteen generations jointly.

%% file: tables/textbody_comparison.tex
\begin{table*}[t]\centering\small
\definecolor{bestred}{RGB}{255,0,0}
\definecolor{secondblue}{RGB}{0,0,255}
\caption{\textbf{Comparison with motion generation baselines on OakInk2.}
All methods generate hand motion from body motion, text, and explicit
hand--time assignments, with $K=16$ generations per input.
We report approximate 95\% confidence intervals
across repeated generations at fixed checkpoints. MinMPJPE and
MModality use all sixteen generations jointly and are point estimates.
Color highlights distinguish the best (\textcolor{bestred}{\textbf{red}})
and second-best (\textcolor{secondblue}{\textbf{blue}}) scores.
Diversity ($\rightarrow$) is ranked by proximity to the ground-truth value.
* denotes baselines adapted to our task.}
\label{tab:textbodycomparison}
\setlength{\tabcolsep}{2.6pt}
\renewcommand{\arraystretch}{1.18}
\begin{tabular*}{\textwidth}{@{\extracolsep{\fill}}lrrrrrr@{}}\toprule
Method & \shortstack{MinMPJPE$\downarrow$\\(mm)} & \shortstack{AvgMPJPE$\downarrow$\\(mm)} & \shortstack{Wrist rotation\\error$\downarrow$ ($^\circ$)} & FHD$\downarrow$ & Diversity$\rightarrow$ & MModality$\uparrow$\\\midrule
Ground truth & --- & --- & --- & 0 & 9.397 & ---\\\midrule
MDM* & 31.002 & \ranksecond{\scoreci{33.094}{0.025}} & \ranksecond{\scoreci{18.26}{0.01}} & \ranksecond{\scoreci{0.532}{0.008}} & \scoreci{9.724}{0.003} & 0.683\\
T2M-GPT* & 30.525 & \scoreci{35.450}{0.049} & \scoreci{19.65}{0.03} & \scoreci{0.872}{0.007} & \scoreci{9.159}{0.007} & 1.607\\
MLD* & \ranksecond{25.725} & \scoreci{34.051}{0.057} & \scoreci{18.98}{0.04} & \scoreci{1.226}{0.131} & \ranksecond{\scoreci{9.608}{0.018}} & \ranksecond{2.403}\\
BOTH2Hands* & 29.913 & \scoreci{39.817}{0.112} & \scoreci{22.52}{0.06} & \scoreci{3.189}{0.061} & \scoreci{8.939}{0.017} & 2.337\\
HandX* & 29.239 & \scoreci{34.578}{0.045} & \scoreci{19.11}{0.03} & \scoreci{0.805}{0.008} & \rankfirst{\scoreci{9.315}{0.007}} & 1.663\\
\midrule
ContextHands & \rankfirst{22.809} & \rankfirst{\scoreci{30.711}{0.092}} & \rankfirst{\scoreci{16.93}{0.04}} & \rankfirst{\scoreci{0.276}{0.042}} & \scoreci{9.043}{0.023} & \rankfirst{2.498}\\
\bottomrule\end{tabular*}\end{table*}

%% file: cvpr/05_results.tex
\paperchapter{Results}
\label{sec:results}
\ifdefined\contexthandsreading\else\input{figures/textbody_qualitative_v102}\fi
\input{tables/textbody_controls}
\papersection{Quantitative comparison}
In the primary comparison, all methods receive explicit hand and time assignments
(Table~\ref{tab:textbodycomparison}). ContextHands achieves the lowest hand position error
both for the best of sixteen completions and on average across all sixteen.
It also achieves the lowest wrist rotation error and FHD. For the same body motion
and text, its Multimodality score indicates more variation than MDM and a similar
level of variation to MLD.

With the sampling settings used in the primary comparison, ContextHands generates
five seconds of hand motion (150 frames) in 2.83 seconds on a single RTX~4090,
excluding frozen text encoding. MDM and HandX,
the two baselines closest to ContextHands in FHD, both take longer per sequence.
MLD and T2M-GPT are the only faster methods, and ContextHands achieves lower
hand position errors and FHD than either (Table~\ref{tab:textbodycomparison}).

\papersection{Ablation studies}
\paragraph{Hand Context Encoder.}
The full model and the direct-text variant both retain JQKE
(Table~\ref{tab:textbodycontrols}). The direct-text variant replaces HCE with
cross-attention from each hand-joint token to the text features assigned to its
hand and frame. Learning hand context with HCE yields lower hand position errors
and FHD. Removing the auxiliary hand-prediction loss from the full model narrows
this gap.

Removing the body input to HCE also increases hand position errors and FHD,
even though JQKE continues to provide the generator with body conditions.
This result supports using body motion to learn hand context as well as to
provide joint-specific kinematic cues.

The full HCE yields lower average and best-of-sixteen hand position errors than the
deterministic variant. Both use diffusion sampling. Compared with equal weighting of the
prior- and posterior-conditioned hand losses, adaptive weighting yields lower FHD
and best-of-sixteen position error, but higher average position error.

\paragraph{Joint-Query Kinematic Encoder.}
The part and temporal-scale summaries complement the individual body-joint
features in JQKE (Table~\ref{tab:textbodycontrols}). When the hand-joint queries
can read the 22 body-joint features but cannot attend directly to the summaries,
both hand position errors and FHD increase.

To test whether the generator benefits from receiving a separate body condition
for each joint, we compute the same 16 joint-specific conditions per hand in
both models. The full model passes each condition to its corresponding joint.
The comparison model averages the 16 conditions at each frame and gives that
average to every joint of the hand. Position errors are nearly identical, but
preserving the joint-specific conditions yields a lower FHD (0.276 vs.\ 0.367).

\paragraph{Training strategy.}
Can a model trained to generate hand motion from the body alone use text as
effectively when it is introduced later? Prior-First~\cite{shi2026priorfirst}
learns a body--hand prior before adding semantic conditioning. Motivated by this
approach, we compare joint training with two staged strategies in ContextHands
(Table~\ref{tab:trainingstrategy}).
Starting from the same body-only checkpoint, we either update only the text
input module or also fine-tune the motion model. Fine-tuning the motion model reduces hand
position errors and FHD relative to updating the text module alone. Across the three
strategies evaluated, joint training with body motion and text from the outset
yields the lowest hand position errors and FHD.

\begin{table}[t]\centering\small
\caption{\textbf{Comparison of joint and staged training.}
Both staged variants start from the same body-only checkpoint with
equal additional training budgets. They update either the text input
module alone or the motion model as well. Best values are bold.}
\label{tab:trainingstrategy}
\setlength{\tabcolsep}{3pt}
\begin{tabular}{@{}lrrr@{}}\toprule
Training strategy & \shortstack{MinMPJPE$\downarrow$\\(mm)} & \shortstack{AvgMPJPE$\downarrow$\\(mm)} & FHD$\downarrow$\\\midrule
Joint (ContextHands) & \textbf{22.809} & \textbf{30.711} & \textbf{0.276}\\
\shortstack[l]{Body-only, then\\text module only} & 25.331 & 37.340 & 0.991\\
\shortstack[l]{Body-only, then\\full fine-tuning} & 24.697 & 35.859 & 0.697\\
\bottomrule\end{tabular}\end{table}

\paragraph{Text conditioning at inference.}
We keep the trained ContextHands model and the input body motion fixed.
Removing or shuffling the descriptions increases average hand position error.

\papersection{Qualitative comparison}
\ifdefined\contexthandsreading\input{figures/textbody_qualitative_v102}\fi
Figure~\ref{fig:qualitativetemporal} compares the generated hand poses with the reference.
In both lid grasping (a) and cutting (b), ContextHands reproduces the extended index finger
and the flexion of the neighboring fingers. MLD also captures the index extension in (a), but does not maintain
it across the displayed frames, while MDM extends a different finger
from the reference in (c).

For each object view, we fit the object to the reference hand and use the same object pose
and rendering procedure for all methods. ContextHands more closely matches the reference
finger configuration around the knife handle in (b) and the rod in (d).
The more open configurations of MLD in (b) and BOTH2Hands in (d) leave
larger visible gaps around the displayed objects. These wrist-aligned
views show finger configuration. Table~\ref{tab:textbodycomparison} reports
wrist rotation error separately.

%% file: figures/textbody_qualitative_v102.tex
\ifdefined\contexthandsreading\begin{figure}[tbp]\else\begin{figure*}[t]\fi
\centering
\includegraphics[width=\textwidth]{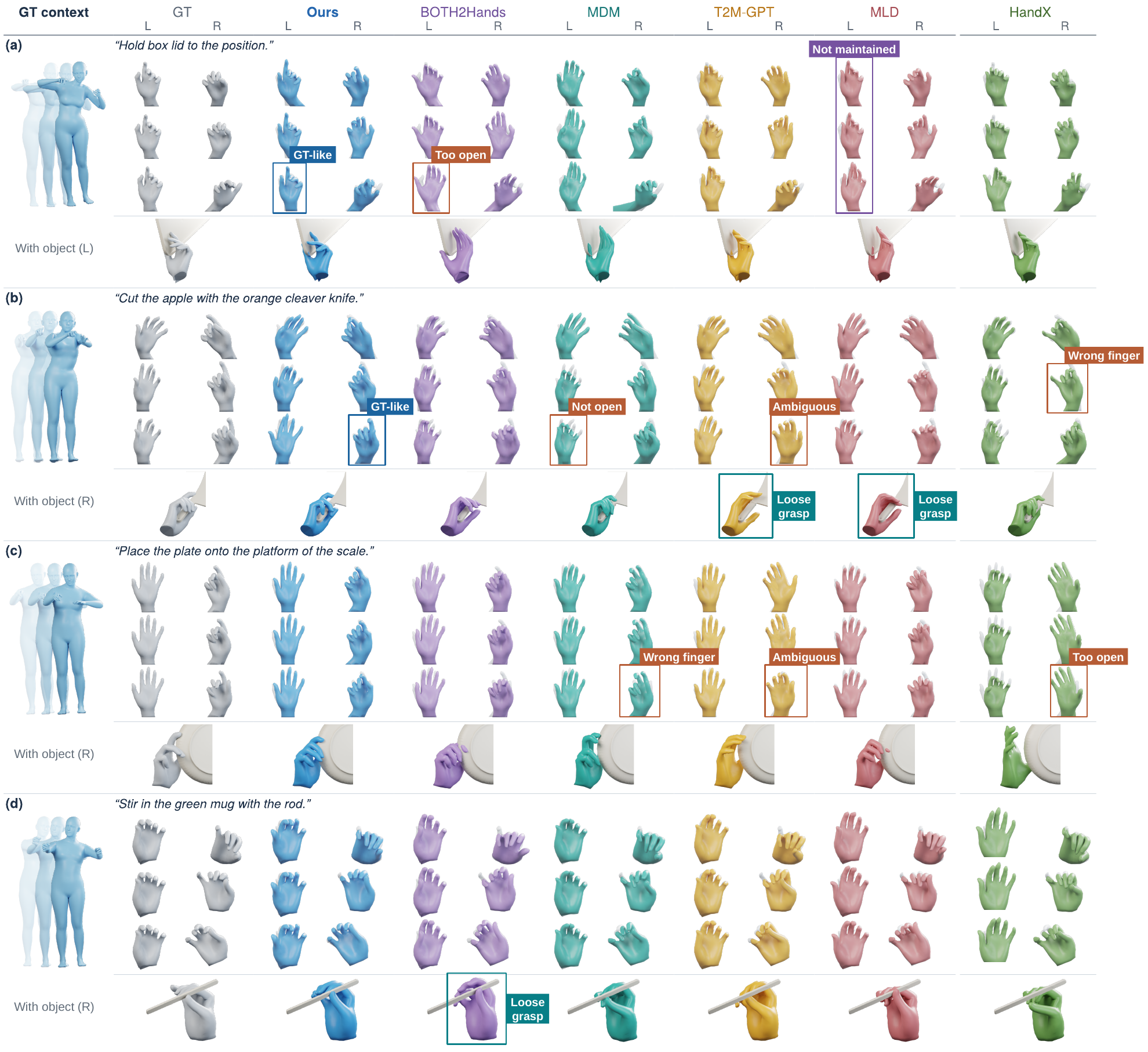}
\caption{Qualitative comparison on OakInk2: (a) grasping a box lid,
(b) cutting an apple, (c) placing a plate on a scale, and (d) stirring with a rod.
Each example uses the same test clip, generation seed, and frame indices
across methods. The three hand poses are ordered in time from top to bottom;
L/R denote the left and right hands. To facilitate comparison of finger
articulation, hands are aligned to the GT wrist position and orientation,
with shared cameras and scale. Translucent GT overlays show differences
in finger poses. Boxes and labels highlight selected details; the annotations are not exhaustive. Object views illustrate grasp shapes using
the same alignment and rendering procedure for GT and all methods.
}
\label{fig:qualitativetemporal}
\ifdefined\contexthandsreading\end{figure}\else\end{figure*}\fi

%% file: tables/textbody_controls.tex
\begin{table}[t]\centering\small
\caption{\textbf{Ablation study of ContextHands.}
All variants use the same denoising backbone. Hand
generation remains stochastic even with Deterministic HCE. Best values are bold.}
\label{tab:textbodycontrols}
\setlength{\tabcolsep}{3pt}
\renewcommand{\arraystretch}{1.18}
\begin{tabular*}{\linewidth}{@{\extracolsep{\fill}}lrrr@{}}\toprule
Variant & \shortstack{MinMPJPE$\downarrow$\\(mm)} & \shortstack{AvgMPJPE$\downarrow$\\(mm)} & FHD$\downarrow$\\\midrule
ContextHands (full) & 22.809 & \textbf{30.711} & \textbf{0.276}\\
\midrule
\multicolumn{4}{@{}l}{\textit{HCE: Learning hand context}}\\
w/o HCE (direct text) & 26.360 & 39.176 & 1.508\\
w/o auxiliary hand loss & 25.930 & 38.545 & 1.295\\
w/o body input to HCE & 23.140 & 32.781 & 0.475\\
Deterministic HCE & 25.577 & 37.092 & 1.127\\
\midrule
\multicolumn{4}{@{}l}{\textit{JQKE: Joint-specific body conditioning}}\\
{\renewcommand{\arraystretch}{1.0}\begin{tabular}[c]{@{}l@{}}Within-hand shared\\body conditioning\end{tabular}} & \textbf{22.785} & 30.818 & 0.367\\
{\renewcommand{\arraystretch}{1.0}\begin{tabular}[c]{@{}l@{}}w/o summary-token access\\in JQKE\end{tabular}} & 23.266 & 32.663 & 0.554\\
\bottomrule\end{tabular*}\end{table}

%% file: cvpr/07_discussion.tex
\paperchapter{Discussion and Limitations}
\label{sec:discussion}
Our results support using body motion both to learn the context of each hand's action
and to guide the motion of individual hand joints. Even when JQKE provided the
generator with body information, including body motion alongside text in HCE reduced
average hand position error. The benefit of joint-specific conditioning was most
apparent in the distribution of generated motion. Passing each joint its own body
condition yielded a lower FHD than sharing the average condition across the hand,
with little difference in position error (Table~\ref{tab:textbodycontrols}).
Under the schedules we evaluated, learning both forms of conditioning together with
the generator also yielded lower position errors and FHD than introducing text
after body-only pretraining (Table~\ref{tab:trainingstrategy}).

We evaluated ContextHands on bimanual object manipulation in OakInk2, where each
available description is associated with a hand and a time interval. The model can
complete intervals without text, but it assumes that these assignments are known
whenever a description is supplied. It also uses future body frames, which limits its
current use to offline completion. Evaluating other activities and descriptions
written freely by users would help establish how well the results generalize. User
studies could further assess whether the generated motions look natural and convey
the actions described.

ContextHands can reproduce the hand pose at the beginning and end of an action yet
miss a brief adjustment in between.

ContextHands does not explicitly model object geometry or hand--object
contact. Incorporating visual observations and contact constraints is a
promising direction for extending ContextHands to object-aware hand
motion generation.\par

%% file: cvpr/06_conclusion.tex
\paperchapter{Conclusion}
\label{sec:conclusion}
We introduced ContextHands to generate detailed hand motion during object manipulation
from body motion and text descriptions assigned to specific hands and time intervals.
Our approach uses body motion in two complementary ways. It helps form a context for
each hand together with the descriptions, and it provides kinematic information that
individual hand joints can draw on separately. An auxiliary hand-pose prediction
task ties the learned context to hand articulation. Both conditioning paths are
trained jointly with a single diffusion generator.

On OakInk2, ContextHands achieved lower hand position errors, wrist rotation error,
and Fr\'echet Hand Distance than all five baselines in the primary comparison
(Table~\ref{tab:textbodycomparison}). The improvement was evident both in how closely
the generated hands followed the reference poses and in how well the generated
motions matched the distribution of real hand motion.

Together, these results support the central idea behind ContextHands. Learning the
context of each hand's action while preserving the kinematic cues relevant to its
individual joints enables the model to generate more precise hand motion during
object manipulation.\par

%% file: cvpr/llm_usage_statement.tex
\section*{LLM Usage Statement}
We used ChatGPT and OpenAI Codex to assist with code development and debugging,
manuscript revision for clarity and readability, and the preparation of draft
visualizations. These tools also assisted with automating and monitoring
pre-specified experiments. The authors developed the core research ideas and
methodology, designed the experiments, and interpreted the results. All
AI-assisted text, code, and figures were manually reviewed and revised as needed
by the authors, who take full responsibility for the accuracy and integrity of
this work.

%% file: main.bbl
\begin{thebibliography}{50}
\providecommand{\natexlab}[1]{#1}
\providecommand{\url}[1]{\texttt{#1}}
\expandafter\ifx\csname urlstyle\endcsname\relax
  \providecommand{\doi}[1]{doi: #1}\else
  \providecommand{\doi}{doi: \begingroup \urlstyle{rm}\Url}\fi

\bibitem[Athanasiou et~al.(2022)Athanasiou, Petrovich, Black, and
  Varol]{athanasiou2022teach}
Nikos Athanasiou, Mathis Petrovich, Michael~J. Black, and G{\"u}l Varol.
\newblock {TEACH}: Temporal action composition for {3D} humans.
\newblock In \emph{International Conference on 3D Vision}, pages 414--423,
  2022.

\bibitem[Athanasiou et~al.(2023)Athanasiou, Petrovich, Black, and
  Varol]{athanasiou2023sinc}
Nikos Athanasiou, Mathis Petrovich, Michael~J. Black, and G{\"u}l Varol.
\newblock {SINC}: Spatial composition of {3D} human motions for simultaneous
  action generation.
\newblock In \emph{Proceedings of the IEEE/CVF International Conference on
  Computer Vision}, pages 9984--9995, 2023.

\bibitem[Bensabath et~al.(2026)Bensabath, Petrovich, and
  Varol]{bensabath2026handmdm}
L{\'e}ore Bensabath, Mathis Petrovich, and G{\"u}l Varol.
\newblock Text-driven {3D} hand motion generation from sign language data.
\newblock In \emph{Proceedings of the IEEE/CVF Conference on Computer Vision
  and Pattern Recognition}, pages 23095--23105, 2026.

\bibitem[Cha et~al.(2024)Cha, Kim, Yoon, and Baek]{cha2024text2hoi}
Junuk Cha, Jihyeon Kim, Jae~Shin Yoon, and Seungryul Baek.
\newblock {Text2HOI}: Text-guided {3D} motion generation for hand-object
  interaction.
\newblock In \emph{Proceedings of the IEEE/CVF Conference on Computer Vision
  and Pattern Recognition}, pages 1577--1585, 2024.

\bibitem[Chen et~al.(2023)Chen, Jiang, Liu, Huang, Fu, Chen, and
  Yu]{chen2023mld}
Xin Chen, Biao Jiang, Wen Liu, Zilong Huang, Bin Fu, Tao Chen, and Gang Yu.
\newblock Executing your commands via motion diffusion in latent space.
\newblock In \emph{Proceedings of the IEEE/CVF Conference on Computer Vision
  and Pattern Recognition}, pages 18000--18010, 2023.

\bibitem[Cheng et~al.(2026)Cheng, Zhu, and He]{cheng2026tshamo}
Ching-Lam Cheng, Bin Zhu, and Shengfeng He.
\newblock Teacher-student diffusion model for text-driven {3D} hand motion
  generation.
\newblock In \emph{IEEE International Conference on Acoustics, Speech and
  Signal Processing}, pages 11972--11976, 2026.

\bibitem[Christen et~al.(2024)Christen, Hampali, Sener, Remelli, Hodan, Sauser,
  Ma, and Tekin]{christen2024diffh2o}
Sammy Christen, Shreyas Hampali, Fadime Sener, Edoardo Remelli, Tomas Hodan,
  Eric Sauser, Shugao Ma, and Bugra Tekin.
\newblock {DiffH2O}: Diffusion-based synthesis of hand-object interactions from
  textual descriptions.
\newblock In \emph{SIGGRAPH Asia 2024 Conference Papers}, pages 1--11, 2024.

\bibitem[Delmas et~al.(2022)Delmas, Weinzaepfel, Lucas, Moreno-Noguer, and
  Rogez]{delmas2022posescript}
Ginger Delmas, Philippe Weinzaepfel, Thomas Lucas, Francesc Moreno-Noguer, and
  Gr{\'e}gory Rogez.
\newblock {PoseScript}: {3D} human poses from natural language.
\newblock In \emph{European Conference on Computer Vision}, pages 346--362,
  2022.

\bibitem[Ghosh et~al.(2023)Ghosh, Dabral, Golyanik, Theobalt, and
  Slusallek]{ghosh2023imos}
Anindita Ghosh, Rishabh Dabral, Vladislav Golyanik, Christian Theobalt, and
  Philipp Slusallek.
\newblock {IMoS}: Intent-driven full-body motion synthesis for human-object
  interactions.
\newblock \emph{Computer Graphics Forum}, 42\penalty0 (2):\penalty0 1--12,
  2023.

\bibitem[Guo et~al.(2024)Guo, Mu, Javed, Wang, and Cheng]{guo2024momask}
Chuan Guo, Yuxuan Mu, Muhammad~Gohar Javed, Sen Wang, and Li Cheng.
\newblock {MoMask}: Generative masked modeling of {3D} human motions.
\newblock In \emph{Proceedings of the IEEE/CVF Conference on Computer Vision
  and Pattern Recognition}, pages 1900--1910, 2024.

\bibitem[He et~al.(2026)He, Tiwari, Zhang, Bora, Birdal, Lenssen, and
  Pons-Moll]{he2026molingo}
Yannan He, Garvita Tiwari, Xiaohan Zhang, Pankaj Bora, Tolga Birdal, Jan~Eric
  Lenssen, and Gerard Pons-Moll.
\newblock {MoLingo}: Motion--language alignment for text-to-human motion
  generation.
\newblock In \emph{Proceedings of the IEEE/CVF Conference on Computer Vision
  and Pattern Recognition}, pages 38387--38398, 2026.

\bibitem[Heusel et~al.(2017)Heusel, Ramsauer, Unterthiner, Nessler, and
  Hochreiter]{heusel2017fid}
Martin Heusel, Hubert Ramsauer, Thomas Unterthiner, Bernhard Nessler, and Sepp
  Hochreiter.
\newblock {GANs} trained by a two time-scale update rule converge to a local
  {Nash} equilibrium.
\newblock In \emph{Advances in Neural Information Processing Systems}, 2017.

\bibitem[Ho et~al.(2020)Ho, Jain, and Abbeel]{ho2020ddpm}
Jonathan Ho, Ajay Jain, and Pieter Abbeel.
\newblock Denoising diffusion probabilistic models.
\newblock In \emph{Advances in Neural Information Processing Systems}, pages
  6840--6851, 2020.

\bibitem[Jiang et~al.(2023)Jiang, Chen, Liu, Yu, Yu, and
  Chen]{jiang2023motiongpt}
Biao Jiang, Xin Chen, Wen Liu, Jingyi Yu, Gang Yu, and Tao Chen.
\newblock {MotionGPT}: Human motion as a foreign language.
\newblock In \emph{Advances in Neural Information Processing Systems}, pages
  20067--20079, 2023.

\bibitem[Karunratanakul et~al.(2023)Karunratanakul, Preechakul, Suwajanakorn,
  and Tang]{karunratanakul2023gmd}
Korrawe Karunratanakul, Konpat Preechakul, Supasorn Suwajanakorn, and Siyu
  Tang.
\newblock Guided motion diffusion for controllable human motion synthesis.
\newblock In \emph{Proceedings of the IEEE/CVF International Conference on
  Computer Vision}, pages 2151--2162, 2023.

\bibitem[Lee et~al.(2024)Lee, Saito, Nam, Sung, and Kim]{lee2024interhandgen}
Jihyun Lee, Shunsuke Saito, Giljoo Nam, Minhyuk Sung, and Tae-Kyun Kim.
\newblock {InterHandGen}: Two-hand interaction generation via cascaded reverse
  diffusion.
\newblock In \emph{Proceedings of the IEEE/CVF Conference on Computer Vision
  and Pattern Recognition}, pages 527--537, 2024.

\bibitem[Lin(2025)]{lin2025handdiffuse}
Pei Lin.
\newblock {HandDiffuse}: Generative controllers for two-hand interactions via
  diffusion models.
\newblock \emph{Proceedings of the AAAI Conference on Artificial Intelligence},
  39\penalty0 (5):\penalty0 5280--5288, 2025.

\bibitem[Liu et~al.(2024)Liu, Zhu, Becherini, Peng, Su, Zhou, Zhe, Iwamoto,
  Zheng, and Black]{liu2024emage}
Haiyang Liu, Zihao Zhu, Giorgio Becherini, Yichen Peng, Mingyang Su, You Zhou,
  Xuefei Zhe, Naoya Iwamoto, Bo Zheng, and Michael~J. Black.
\newblock {EMAGE}: Towards unified holistic co-speech gesture generation via
  expressive masked audio gesture modeling.
\newblock In \emph{Proceedings of the IEEE/CVF Conference on Computer Vision
  and Pattern Recognition}, pages 1144--1154, 2024.

\bibitem[Loshchilov and Hutter(2019)]{loshchilov2019adamw}
Ilya Loshchilov and Frank Hutter.
\newblock Decoupled weight decay regularization.
\newblock In \emph{International Conference on Learning Representations}, 2019.

\bibitem[Lu et~al.(2024)Lu, Chen, Zeng, Lin, Zhang, Zhang, and
  Shum]{lu2024humantomato}
Shunlin Lu, Ling-Hao Chen, Ailing Zeng, Jing Lin, Ruimao Zhang, Lei Zhang, and
  Heung-Yeung Shum.
\newblock {HumanTOMATO}: Text-aligned whole-body motion generation.
\newblock In \emph{Proceedings of the 41st International Conference on Machine
  Learning}, pages 32939--32977. PMLR, 2024.

\bibitem[Ng et~al.(2021)Ng, Ginosar, Darrell, and Joo]{ng2021body2hands}
Evonne Ng, Shiry Ginosar, Trevor Darrell, and Hanbyul Joo.
\newblock {Body2Hands}: Learning to infer {3D} hands from conversational
  gesture body dynamics.
\newblock In \emph{Proceedings of the IEEE/CVF Conference on Computer Vision
  and Pattern Recognition}, pages 11865--11874, 2021.

\bibitem[Nichol and Dhariwal(2021)]{nichol2021improved}
Alexander~Quinn Nichol and Prafulla Dhariwal.
\newblock Improved denoising diffusion probabilistic models.
\newblock In \emph{Proceedings of the 38th International Conference on Machine
  Learning}, pages 8162--8171. PMLR, 2021.

\bibitem[Pavlakos et~al.(2019)Pavlakos, Choutas, Ghorbani, Bolkart, Osman,
  Tzionas, and Black]{pavlakos2019smplx}
Georgios Pavlakos, Vasileios Choutas, Nima Ghorbani, Timo Bolkart, Ahmed A.~A.
  Osman, Dimitrios Tzionas, and Michael~J. Black.
\newblock Expressive body capture: {3D} hands, face, and body from a single
  image.
\newblock In \emph{Proceedings of the IEEE/CVF Conference on Computer Vision
  and Pattern Recognition}, pages 10975--10985, 2019.

\bibitem[Petrovich et~al.(2022)Petrovich, Black, and Varol]{petrovich2022temos}
Mathis Petrovich, Michael~J. Black, and G{\"u}l Varol.
\newblock {TEMOS}: Generating diverse human motions from textual descriptions.
\newblock In \emph{European Conference on Computer Vision}, pages 480--497,
  2022.

\bibitem[Petrovich et~al.(2023)Petrovich, Black, and Varol]{petrovich2023tmr}
Mathis Petrovich, Michael~J. Black, and G{\"u}l Varol.
\newblock {TMR}: Text-to-motion retrieval using contrastive {3D} human motion
  synthesis.
\newblock In \emph{Proceedings of the IEEE/CVF International Conference on
  Computer Vision}, pages 9488--9497, 2023.

\bibitem[Petrovich et~al.(2024)Petrovich, Litany, Iqbal, Black, Varol, Peng,
  and Rempe]{petrovich2024stmc}
Mathis Petrovich, Or Litany, Umar Iqbal, Michael~J. Black, G{\"u}l Varol,
  Xue~Bin Peng, and Davis Rempe.
\newblock Multi-track timeline control for text-driven {3D} human motion
  generation.
\newblock In \emph{Proceedings of the IEEE/CVF Conference on Computer Vision
  and Pattern Recognition Workshops}, pages 1911--1921, 2024.

\bibitem[Pinyoanuntapong et~al.(2024)Pinyoanuntapong, Wang, Lee, and
  Chen]{pinyoanuntapong2024mmm}
Ekkasit Pinyoanuntapong, Pu Wang, Minwoo Lee, and Chen Chen.
\newblock {MMM}: Generative masked motion model.
\newblock In \emph{Proceedings of the IEEE/CVF Conference on Computer Vision
  and Pattern Recognition}, pages 1546--1555, 2024.

\bibitem[Qi et~al.(2023)Qi, Liu, Sun, Li, Fan, and Yu]{qi2023bhd}
Xingqun Qi, Chen Liu, Muyi Sun, Lincheng Li, Changjie Fan, and Xin Yu.
\newblock Diverse {3D} hand gesture prediction from body dynamics by bilateral
  hand disentanglement.
\newblock In \emph{Proceedings of the IEEE/CVF Conference on Computer Vision
  and Pattern Recognition}, pages 4616--4626, 2023.

\bibitem[Radford et~al.(2021)Radford, Kim, Hallacy, Ramesh, Goh, Agarwal,
  Sastry, Askell, Mishkin, Clark, Krueger, and Sutskever]{radford2021clip}
Alec Radford, Jong~Wook Kim, Chris Hallacy, Aditya Ramesh, Gabriel Goh,
  Sandhini Agarwal, Girish Sastry, Amanda Askell, Pamela Mishkin, Jack Clark,
  Gretchen Krueger, and Ilya Sutskever.
\newblock Learning transferable visual models from natural language
  supervision.
\newblock In \emph{Proceedings of the 38th International Conference on Machine
  Learning}, pages 8748--8763. PMLR, 2021.

\bibitem[Raffel et~al.(2020)Raffel, Shazeer, Roberts, Lee, Narang, Matena,
  Zhou, Li, and Liu]{raffel2020t5}
Colin Raffel, Noam Shazeer, Adam Roberts, Katherine Lee, Sharan Narang, Michael
  Matena, Yanqi Zhou, Wei Li, and Peter~J. Liu.
\newblock Exploring the limits of transfer learning with a unified text-to-text
  transformer.
\newblock \emph{Journal of Machine Learning Research}, 21\penalty0
  (140):\penalty0 1--67, 2020.

\bibitem[Ren et~al.(2018)Ren, Zeng, Yang, and Urtasun]{ren2018reweight}
Mengye Ren, Wenyuan Zeng, Bin Yang, and Raquel Urtasun.
\newblock Learning to reweight examples for robust deep learning.
\newblock In \emph{Proceedings of the 35th International Conference on Machine
  Learning}, pages 4334--4343. PMLR, 2018.

\bibitem[Sanh et~al.(2019)Sanh, Debut, Chaumond, and Wolf]{sanh2019distilbert}
Victor Sanh, Lysandre Debut, Julien Chaumond, and Thomas Wolf.
\newblock {DistilBERT}, a distilled version of {BERT}: smaller, faster, cheaper
  and lighter.
\newblock \emph{arXiv preprint arXiv:1910.01108}, 2019.

\bibitem[Shi et~al.(2026)Shi, Chen, and Komura]{shi2026priorfirst}
Mingyi Shi, Xuelin Chen, and Taku Komura.
\newblock Prior-first, condition-second: Scalable and controllable hand motion
  completion.
\newblock \emph{Computer Graphics Forum}, 45\penalty0 (8), 2026.

\bibitem[Sohn et~al.(2015)Sohn, Yan, and Lee]{sohn2015cvae}
Kihyuk Sohn, Xinchen Yan, and Honglak Lee.
\newblock Learning structured output representation using deep conditional
  generative models.
\newblock In \emph{Advances in Neural Information Processing Systems}, 2015.

\bibitem[Tevet et~al.(2022)Tevet, Gordon, Hertz, Bermano, and
  Cohen-Or]{tevet2022motionclip}
Guy Tevet, Brian Gordon, Amir Hertz, Amit~H. Bermano, and Daniel Cohen-Or.
\newblock {MotionCLIP}: Exposing human motion generation to {CLIP} space.
\newblock In \emph{European Conference on Computer Vision}, pages 358--374,
  2022.

\bibitem[Tevet et~al.(2023)Tevet, Raab, Gordon, Shafir, Cohen-Or, and
  Bermano]{tevet2023mdm}
Guy Tevet, Sigal Raab, Brian Gordon, Yonatan Shafir, Daniel Cohen-Or, and
  Amit~H. Bermano.
\newblock Human motion diffusion model.
\newblock In \emph{International Conference on Learning Representations}, 2023.

\bibitem[Thambiraja et~al.(2026)Thambiraja, Taheri, Danecek, Becherini,
  Pons-Moll, and Thies]{thambiraja2026clutch}
Balamurugan Thambiraja, Omid Taheri, Radek Danecek, Giorgio Becherini, Gerard
  Pons-Moll, and Justus Thies.
\newblock {CLUTCH}: Contextualized language model for unlocking
  text-conditioned hand motion modelling in the wild.
\newblock In \emph{International Conference on Learning Representations}, 2026.

\bibitem[Vaswani et~al.(2017)Vaswani, Shazeer, Parmar, Uszkoreit, Jones, Gomez,
  Kaiser, and Polosukhin]{vaswani2017attention}
Ashish Vaswani, Noam Shazeer, Niki Parmar, Jakob Uszkoreit, Llion Jones,
  Aidan~N. Gomez, {\L}ukasz Kaiser, and Illia Polosukhin.
\newblock Attention is all you need.
\newblock In \emph{Advances in Neural Information Processing Systems}, 2017.

\bibitem[Xie et~al.(2024)Xie, Jampani, Zhong, Sun, and
  Jiang]{xie2024omnicontrol}
Yiming Xie, Varun Jampani, Lei Zhong, Deqing Sun, and Huaizu Jiang.
\newblock {OmniControl}: Control any joint at any time for human motion
  generation.
\newblock In \emph{International Conference on Learning Representations}, 2024.

\bibitem[Yi et~al.(2023)Yi, Liang, Liu, Cao, Wen, Bolkart, Tao, and
  Black]{yi2023talkshow}
Hongwei Yi, Hualin Liang, Yifei Liu, Qiong Cao, Yandong Wen, Timo Bolkart,
  Dacheng Tao, and Michael~J. Black.
\newblock Generating holistic {3D} human motion from speech.
\newblock In \emph{Proceedings of the IEEE/CVF Conference on Computer Vision
  and Pattern Recognition}, pages 469--480, 2023.

\bibitem[Yoon et~al.(2020)Yoon, Cha, Lee, Jang, Lee, Kim, and
  Lee]{yoon2020speech}
Youngwoo Yoon, Bok Cha, Joo-Haeng Lee, Minsu Jang, Jaeyeon Lee, Jaehong Kim,
  and Geehyuk Lee.
\newblock Speech gesture generation from the trimodal context of text, audio,
  and speaker identity.
\newblock \emph{ACM Transactions on Graphics}, 39\penalty0 (6):\penalty0 1--16,
  2020.
\newblock Article 222.

\bibitem[Zhan et~al.(2024)Zhan, Yang, Zhao, Mao, Xu, Lin, Li, and
  Lu]{zhan2024oakink2}
Xinyu Zhan, Lixin Yang, Yifei Zhao, Kangrui Mao, Hanlin Xu, Zenan Lin, Kailin
  Li, and Cewu Lu.
\newblock {OAKINK2}: A dataset of bimanual hands-object manipulation in complex
  task completion.
\newblock In \emph{Proceedings of the IEEE/CVF Conference on Computer Vision
  and Pattern Recognition}, pages 445--456, 2024.

\bibitem[Zhang et~al.(2023{\natexlab{a}})Zhang, Zhang, Cun, Zhang, Zhao, Lu,
  Shen, and Shan]{zhang2023t2mgpt}
Jianrong Zhang, Yangsong Zhang, Xiaodong Cun, Yong Zhang, Hongwei Zhao, Hongtao
  Lu, Xi Shen, and Ying Shan.
\newblock Generating human motion from textual descriptions with discrete
  representations.
\newblock In \emph{Proceedings of the IEEE/CVF Conference on Computer Vision
  and Pattern Recognition}, pages 14730--14740, 2023{\natexlab{a}}.

\bibitem[Zhang et~al.(2023{\natexlab{b}})Zhang, Guo, Pan, Cai, Hong, Li, Yang,
  and Liu]{zhang2023remodiffuse}
Mingyuan Zhang, Xinying Guo, Liang Pan, Zhongang Cai, Fangzhou Hong, Huirong
  Li, Lei Yang, and Ziwei Liu.
\newblock {ReMoDiffuse}: Retrieval-augmented motion diffusion model.
\newblock In \emph{Proceedings of the IEEE/CVF International Conference on
  Computer Vision}, pages 364--373, 2023{\natexlab{b}}.

\bibitem[Zhang et~al.(2023{\natexlab{c}})Zhang, Li, Cai, Ren, Yang, and
  Liu]{zhang2023finemogen}
Mingyuan Zhang, Huirong Li, Zhongang Cai, Jiawei Ren, Lei Yang, and Ziwei Liu.
\newblock {FineMoGen}: Fine-grained spatio-temporal motion generation and
  editing.
\newblock In \emph{Advances in Neural Information Processing Systems}, pages
  13981--13992, 2023{\natexlab{c}}.

\bibitem[Zhang et~al.(2024{\natexlab{a}})Zhang, Cai, Pan, Hong, Guo, Yang, and
  Liu]{zhang2024motiondiffuse}
Mingyuan Zhang, Zhongang Cai, Liang Pan, Fangzhou Hong, Xinying Guo, Lei Yang,
  and Ziwei Liu.
\newblock {MotionDiffuse}: Text-driven human motion generation with diffusion
  model.
\newblock \emph{IEEE Transactions on Pattern Analysis and Machine
  Intelligence}, 46\penalty0 (6):\penalty0 4115--4128, 2024{\natexlab{a}}.

\bibitem[Zhang et~al.(2025)Zhang, Liu, Garrido, Kim, and
  Chaudhuri]{zhang2025kinmo}
Pengfei Zhang, Pinxin Liu, Pablo Garrido, Hyeongwoo Kim, and Bindita Chaudhuri.
\newblock {KinMo}: Kinematic-aware human motion understanding and generation.
\newblock In \emph{Proceedings of the IEEE/CVF International Conference on
  Computer Vision}, pages 11187--11197, 2025.

\bibitem[Zhang et~al.(2024{\natexlab{b}})Zhang, Huang, Zhou, Zhang, Yu, Wang,
  and Xu]{zhang2024both2hands}
Wenqian Zhang, Molin Huang, Yuxuan Zhou, Juze Zhang, Jingyi Yu, Jingya Wang,
  and Lan Xu.
\newblock {BOTH2Hands}: Inferring {3D} hands from both text prompts and body
  dynamics.
\newblock In \emph{Proceedings of the IEEE/CVF Conference on Computer Vision
  and Pattern Recognition}, pages 2393--2404, 2024{\natexlab{b}}.

\bibitem[Zhang et~al.(2026)Zhang, Zhang, Xu, Wang, Xu, Zhou, Zhou, Guo, Wang,
  Wang, and Gui]{zhang2026handx}
Zimu Zhang, Yucheng Zhang, Xiyan Xu, Ziyin Wang, Sirui Xu, Kai Zhou, Bing Zhou,
  Chuan Guo, Jian Wang, Yu-Xiong Wang, and Liang-Yan Gui.
\newblock {HandX}: Scaling bimanual motion and interaction generation.
\newblock In \emph{Proceedings of the IEEE/CVF Conference on Computer Vision
  and Pattern Recognition}, pages 2274--2284, 2026.

\bibitem[Zhou et~al.(2019)Zhou, Barnes, Lu, Yang, and Li]{zhou2019rotation}
Yi Zhou, Connelly Barnes, Jingwan Lu, Jimei Yang, and Hao Li.
\newblock On the continuity of rotation representations in neural networks.
\newblock In \emph{Proceedings of the IEEE/CVF Conference on Computer Vision
  and Pattern Recognition}, pages 5745--5753, 2019.

\end{thebibliography}
